\documentclass{article}
\usepackage{lipsum}
\usepackage{mwe}
\usepackage[utf8]{inputenc}
\usepackage{longtable}
\usepackage{blindtext,alltt}
\usepackage{xcolor}
\usepackage{float}
\usepackage{geometry}
\usepackage{csquotes}
\usepackage[utf8]{inputenc}
\usepackage[english]{babel}
\usepackage[
    backend=biber,
    style=numeric,
    citestyle=numeric,
]{biblatex}
\usepackage{titlesec}
\titleformat{\paragraph}
{\normalfont\normalsize\bfseries}{\theparagraph}{1em}{}
\titlespacing*{\paragraph}
{0pt}{3.25ex plus 1ex minus .2ex}{1.5ex plus .2ex}

\title{Marvelous Minified Models}
\author{Rich Harang \& Hillary Sanders}

\begin{document}
\maketitle


\begin{abstract}
This paper concerns itself with the task of taking a large trained neural network and `compressing' it to be smaller by deleting parameters or entire neurons, with minimal decreases in the resulting model accuracy. We compare various methods of parameter and neuron selection: dropout-based neuron damage estimation, neuron merging, absolute-value based selection, random selection, OBD (Optimal Brain Damage). We also compare a variation on the classic OBD method that slightly outperformed all other parameter and neuron selection methods in our tests with substantial pruning, which we call OBD-SD. We compare these methods against quantization of parameters. We also compare these techniques (all applied to a trained neural network), with neural networks trained from scratch (random weight initialization) on various pruned architectures.
Our results are only barely consistent with the Lottery Ticket Hypothesis \cite{lottery}, in that fine-tuning a parameter-pruned model does \textit{slightly} better than retraining a similarly pruned model from scratch with randomly initialized weights. For neuron-level pruning, retraining from scratch did much better in our experiments.
\end{abstract}

\section{Introduction}
There are many ways to make a deep neural network smaller. 
In this paper, we focus on three categories of model size reduction: \textbf{pruning}, \textbf{quantization}, and \textbf{training smaller models from scratch}.
\newline
\newline
\textbf{Quantization} means changing model parameters to lower-precision formats, like changing all 32-bit floating point parameters to 16-bit, which results in file size about half as large.
\newline
\newline
\textbf{Pruning} deals with deleting parameters or groups of parameters (like entire neurons) from a trained model to make it smaller (often followed by a fine-tuning round of training, as done in our experiments). Parameter-level pruning (also called unstructured pruning) prunes individual parameters at a time, whereas neuron-level pruning (also called structured pruning) prunes all parameters associated with a given neuron at once. 
\newline
\newline
To simplify terminology across multiple methods we use the term 'damage' to broadly refer to the undesired impact of removing a node or zeroing a weight on network performance. Different compression methods use different approaches to either estimate damage directly, or rank neurons or weights in order of increasing assumed damage according to some other metric that does not directly evaluate the impact on loss or performance. This `damage' term, when used in the context of directly estimating the damage to loss caused by pruning a parameter is sometimes referred to as `saliency' in other papers.
\newline
\newline
If a single neuron that is fully connected to its previous and subsequent layer is removed (pruned), the two matrices representing both sets of connections each lose a column or row, resulting in a smaller overall model (in memory and in file size). However, removing a single parameter in a fully connected layer generally doesn't reduce a model's memory requirements or file size, because the weight matrices remain the same dimensions. Currently, there is a lack of general support for sparse operations in neural network deployment libraries, but if these become more widespread, parameter-level pruning will more easily lead to significant memory and file-size savings. Without sparse operations support, parameter level pruning only shines when trying to minimize compressed (e.g. zipped) file size.
\newline
\newline
We found that if zipped file size reduction is the goal, parameter removal tends to outperform neuron removal (lower reductions in accuracy given a reduced zipped file size), whereas if uncompressed file size or memory is what you want to reduce, parameter removal has little benefit. To make results comparable, our inline plots show reductions in accuracy with respect to percent reductions in zipped model file size. 

\footnote{Although the methods discussed in this paper would work fine on non-standard neural network layers (like convolutions), we only implemented this method on fully connected layers, mostly due to ease of implementation and ubiquitousness of fully connected layers in modern neural networks.}
\newline
\newline
Finally, we also train various reduced size - in terms of number of neurons - model architectures from scratch: one architecture suggested by our OBD-SD neuron pruning method, as well as various architectures resulting from fixed reductions in fully connected layer sizes. We also train various reduced size - in terms of number of parameters - from scratch resulting from fixed reductions in the number of connections between (originally fully connected) layers.

\section{Background \& Related Work}
While quantization is a fairly straightforward way of reducing model size, model pruning is more complicated. It's not obvious which parts of your model should be removed to cause the least amount of damage.
\newline
\newline
Optimal Brain Damage\cite{obd} attempts to estimate the change in loss that will be caused by setting an individual parameter to 0. While computationally difficult to compute directly, this can be estimated with the help of Taylor Series and a few assumptions. These damage estimates can be aggregated to the neuron level in order to choose which neurons to remove. Various other papers have experimented with OBD and variants of the OBD method\cite{obs-layer-wise}\cite{obs2}\cite{ebd}. All such papers we can find consider use empirical averages to estimate the Hessian for their OBD-related approaches; we introduce an approach (OBD-SD) that focuses on the empirical variance (see Methods section for details).
Dan et al.\cite{absolute} Experiment with parameter (neuron connection) pruning based on the absolute value of parameters; we replicate this approach in our experiments (and extend to the neuron level).
\newline
\newline
Zhong et al.\cite{merging} propose merging similar neurons in the same layers together. This can be extended by prioritizing neuron pairs that are both similar in terms of parameters, and score low in terms of damage estimates.
\newline
\newline
Salehinejad et al.\cite{EDropout} propose using dropout to find subnetworks that don't result in high loss values. We implement a similar approach, where dropout is used as a means of generating data to estimate neuron importance via a linear regression.
\newline
\newline
Liu et al.\cite{rethinking} interestingly concludes that a large model pruned to a smaller size does no better (or does worse) than training a smaller model from scratch, contradicting implications of the Lottery Ticket Hypothesis\cite{lottery}. This suggests that pruning may be useful for architecture search, but does not bring improvements that pruning has generally been thought to achieve. Because of these results, we also compare our pruned models to models of the same resulting architecture, trained from scratch with random weight initialization (with both parameter-level and neuron-level pruning). Our results are consistent with the Lottery Ticket Hypothesis\cite{lottery}, though by a very slim margin, and only for parameter-level pruning. Specifically, we found than fine-tuning a parameter-pruned model with our best methods does slightly better than retraining a similarly sized model from scratch with random initializations. However, we found that fine-tuning a neuron-pruned model does consistently worse than retraining a similarly pruned model from scratch with random initializations.


\section{Methods}
We used the same `base' model in all of our experiments: our portable executable malware detection deep neural network, which is entirely comprised of dense, feed-forward layers (in addition to dropout and batch-norm `layers'). The model is described in detail in Harang et. al \cite{aloha}'s paper, except that ours excludes auxiliary tag outputs. The original base model was trained on a dataset of approximately 20 million samples for 20 epochs. Trained-from-scratch models used the same dataset. Pruning methods with fine-tuning also used this same dataset. Evaluation was performed on the same time-split test set of size 3 million never-before-seen (in our dataset) samples for all results. The main output of the model is the binary `is\_malware' prediction, which had a 75.3\% positive class balance in the training dataset, and a 79.9\% positive class balance in the test dataset. This is the output our accuracy results refer to.
\newline
\newline
In our pruning experiments, we iteratively pruned $p=10$ percent neurons or parameters (neuron connections) from each of the first five fully connected layers in the network (output layers were not pruned) and then fine-tuned the smaller model on a random selection of 3 million samples from the original training dataset (resampled during each fine-tuning pass).

\subsection{Model Compression Methods}

\subsubsection{Pruning Methods}
\begin{enumerate}
    \item \textbf{Parameter-Level Damage Estimation Techniques} These techniques were all also extended to the neuron-level evaluation by summing damage estimates over each neuron's parameters.
    \begin{itemize}
        \item \textbf{Random} \newline
        Parameters in each layer are deleted at random (fixed proportion per layer per pruning round).
        \item \textbf{Magnitude-based} \newline
        Parameters are selected to be deleted by their absolute value, prioritizing the deletion of parameters close to 0. The idea is that it doesn't cause much damage to set parameters close to 0, to 0. The `damage' ranking (where lowest damage parameters are pruned first) $d_i$ for parameter $\theta_i$ is defined as:
            \begin{equation}
                \hat{d}_{i} = \mid \theta_i \mid
            \end{equation}
    
    \item \textbf{Optimal Brain Damage (OBD)} \newline
        Parameters are selected to be deleted by selecting parameters with the smallest OBD\cite{obd} damage estimates.
        \newline
        \newline
        OBD attempts to estimate the change in loss $L$ that would be caused by setting a parameter $\theta_i$ to $\theta_i'=0$, that is: $L(\theta_{i}') - L(\theta_{i})$. OBD approximates $L(\theta')$ by using Taylor Series (as recalculating loss individually for each parameter in a neural network across a sufficient number of test samples would be very computationally expensive):
        
            \begin{equation}
                L(\theta_i') = L(\theta_i) +   \frac{\partial L(\theta_i)}{\partial \theta_i}\theta_i + \frac{1}{2} \frac{\partial^2 L(\theta_i)}{\partial \theta_i^2}\theta_i^2 + ... 
            \end{equation}
        \newline
        \newline 
        OBD assumes away  the higher order terms ($...$) in the equation above, as well as assumes that $\frac{\partial L}{\partial \theta_i}$ is 0, because loss of a trained model is at a local minimum (and OBD is applied to trained models). So the damage estimate comes out to:
            \begin{equation}
                \hat{d}_{i} = \frac{1}{2} \frac{\partial^2 L}{\partial \theta_i^2}\theta_i^2
            \end{equation}
            
        In our experiments, we used a sample from the training dataset to estimate loss. 
    
    \item \textbf{Standard Deviation or Variance Based Optimal Brain Damage for Large Models (OBD-SD)}
        While the theory behind the OBD technique is elegant, in practice we found that there are some stumbling blocks, particularly if the loss function you're approximating with Taylor Series is complex (i.e. a large model with many layers). Taylor Series approximations become less and less accurate the larger the change in a parameter $\theta_i$ is (and since we're setting parameters to 0, this is often quite large), and in general the more complex the function is. Additionally, OBD assumes away non-diagonal Hessian terms.
        \newline
        \newline
        If we trust that a model has been trained to a good local minimum, it would be surprising if setting parameters to 0 would result in a loss reduction. So we expected our damage estimates to be almost all non-negative. However, in practice, we found only about two-thirds of damage estimates to be non-negative. On further inspection, we found that (even with a large sample size of 100,000) there was almost zero correlation between $mean(\frac{\partial^2 L}{\partial \theta_i^2}\theta_i^2)$ and $SD( \frac{\partial^2 L}{\partial \theta_i^2}\theta_i^2)$, but a very strong positive correlation between the absolute of these two variables, around .8-.9 (see Figure 1). 
        \newline
        \newline

        \begin{figure}[htbp]
                \centering
                \includegraphics[width=1\textwidth]{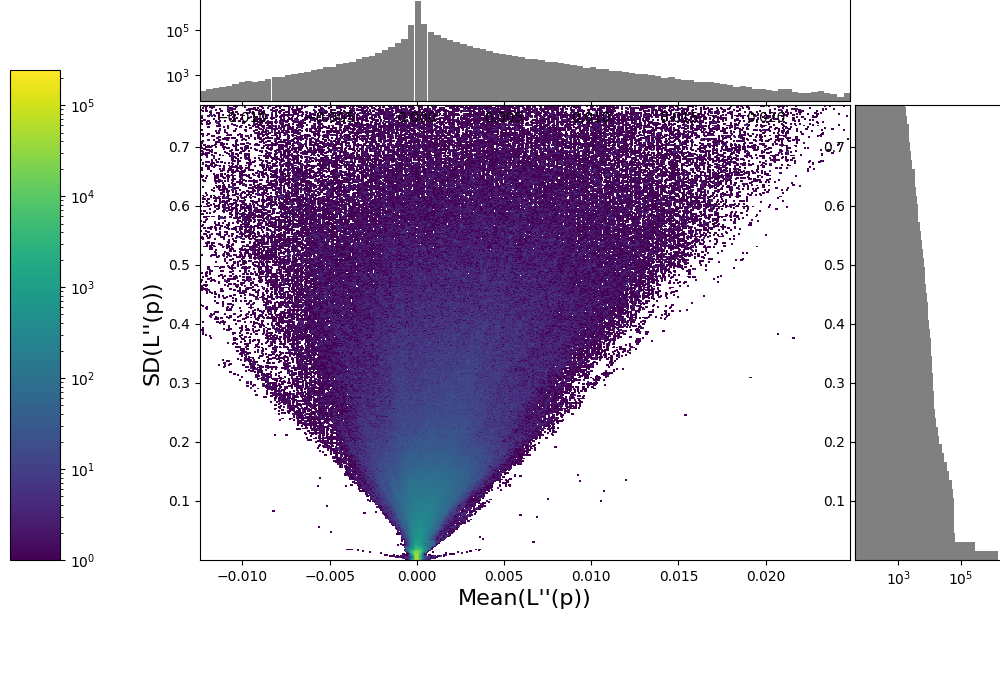}
                \caption{$mean(\frac{\partial^2 L}{\partial \theta_i^2}\theta_i^2)$ plotted against $SD( \frac{\partial^2 L}{\partial \theta_i^2}\theta_i^2)$  across a large sample of model parameters. Note how parameters with large absolute mean scores almost never have very low SD scores (thus the ``V'' shape of the plot). (Note: the sample size used to estimate mean and SDs was 100,000, making the confidence interval of each value quite small.)}
        \end{figure}
        \begin{figure}[htbp]
                \centering
                \includegraphics[width=1\textwidth]{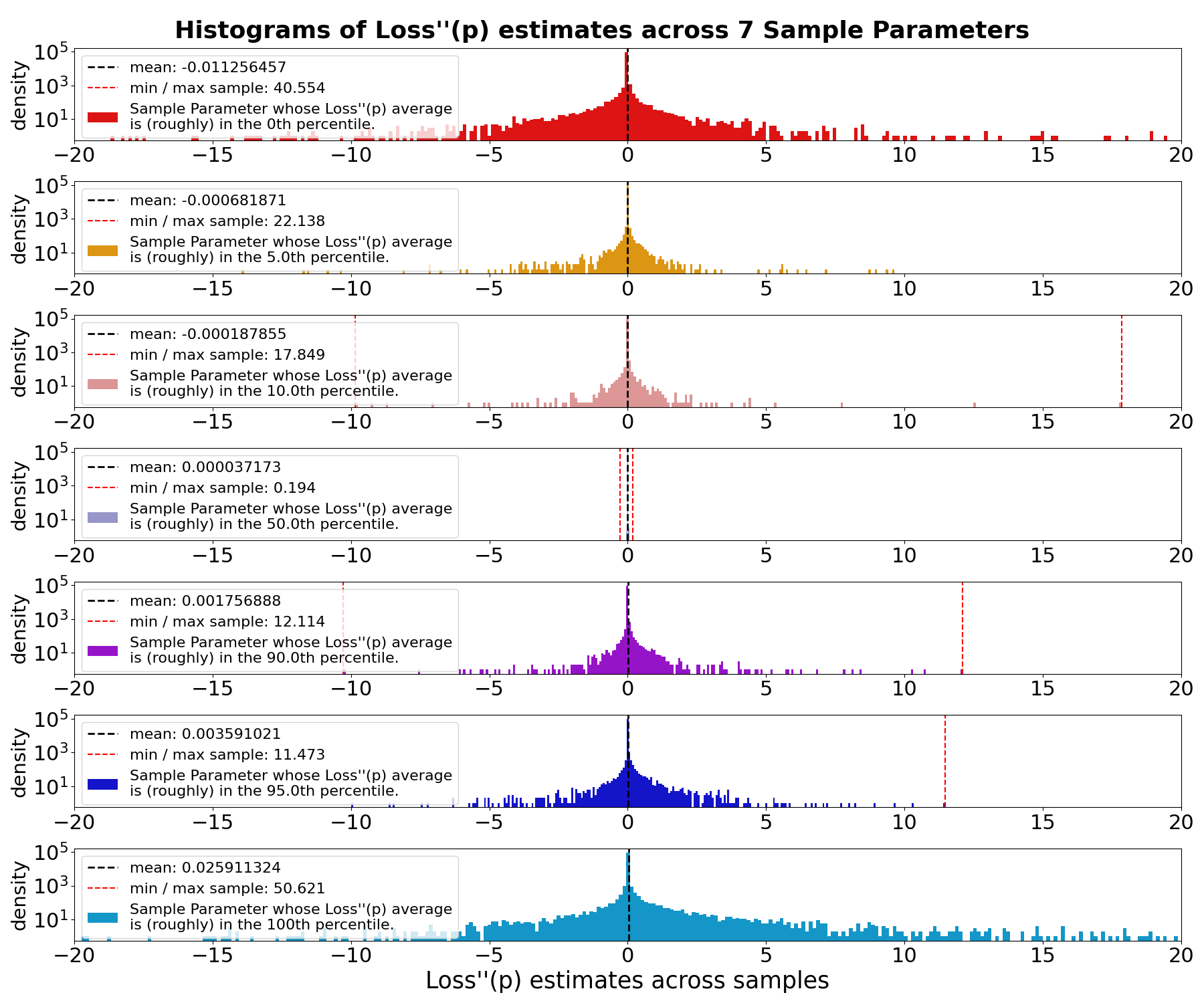}
                \caption{AUC comparison of various parameter pruning methods (using 10\% pruning on fully-connected layers with one fine-tuning epoch each round).}
        \end{figure}

        These results indicated that the OBD damage score is not altogether trustworthy (indeed, it performed worse than random in our results).\footnote{In the original OBD paper (published in 1989) the authors lacked a procedure to directly calculate the hessian diagonal second derivatives $\frac{\partial^2 L}{\partial \theta_i^2}$. They mention the Levenberg-Marquardt approximation, which appears to approximate this value with $2\frac{\partial L}{\partial \theta_i}^2$), which *does* guarantee a positive damage estimate (and, note, is the same as $2SD(\frac{\partial L}{\partial \theta_i})$ when loss is at a local minimum). In our experiments though, this approximation had very little correlation with the thing it was attempting to approximate, so we chose to use the actual $\frac{\partial^2 L}{\partial \theta_i^2}$ form. It is interesting to note, though, that the approximation form performed much better than the non-approximated form, though still worse than our OBD-SD approach.} However, the strong correlation between the absolute value of the mean OBD damage scores and the variance of them indicated that perhaps while the direction of each OBD damage score might be untrustworthy, the overall strength of the scores might be more reliable a signal. Because of this, we decided to explore the more robust signal coming from the variance of the individual sample-level OBD signals. For this reason, we explored ranking parameters by the damage estimate provided by the variance of the OBD damage estimates (change in loss estimates). $Var(\frac{1}{2} \frac{\partial^2 L}{\partial \theta_i^2}\theta_i^2) = \frac{\theta^{4}_i}{4}var(\frac{\partial^2 L}{\partial \theta_i^2})$, which generates the same ordering as the slightly simpler:
        
            \begin{equation}
                \hat{d}_{i} = \theta_i^2 SD(\frac{\partial^2 L}{\partial \theta_i^2})
            \end{equation}
        \newline
        Which we refer to as OBD-SD (Optimal Brain Damage - Standard Deviation) in rest of the paper.
        \end{itemize}
    \item \textbf{Neuron-Level Damage Estimation Techniques}
While all of the parameter-level damage estimation techniques can (and were, in our experiments) be extended to the neuron-level by simply summing damages across all parameters of a given neuron, these methods were only applied to neurons in our experiments.
    \begin{itemize}
        \item \textbf{Dropout-based Pruning}
        OBD and related techniques estimate how loss will be changed when you delete one or more parameters via Taylor Series approximations. Another approach is to calculate the change in loss more directly. Evaluating loss accurately requires running many samples through a neural network, so directly re-evaluating loss for every permutation of your model's architecture (i.e. the current architecture sans one parameter, or sans one neuron) is extremely costly, computationally. However, we hypothesized that you might be able to estimate it by combining the loss results from many dropout rounds (where a random sub-sampling of neurons are deleted) in order to estimate the importance of each neuron.
        \newline
        \newline
        To do this, we calculated loss $L$ from a fixed batch sample over $d$ dropout rounds. We fit a linear regression to these $d$ samples, modeling $L$ as the dependent variable, and boolean values for each neuron (representing whether the neuron was masked or not during dropout) as the independent variables. This provided a computationally cheaper way to estimate the damage caused by deleting any given neuron.
        \newline
        \newline
        \item \textbf{Neuron Merging}
        Computationally, it's quick to calculate the pairwise euclidean distances between each neuron in a network layer, in terms of their parameters. We chose the calculate the distance matrix amongst all neurons in a layer, and then iteratively choose the closest two remaining neurons to merge. We merged neurons by averaging their parameters (weights and biases) and then deleting one of the two neurons. Note that depending on the activation function being used, this technique can have variable effects (as in, the output of two neurons merged into one may not be the same as the average output from the original two neurons). Our neural network used ELU activations in its fully connected layers\footnote{We did run similar tests with a model using RELU activations and noticed no significant differences in the results} \footnote{We also tried combining this approach with damage estimates from other approaches by adding damage estimates into the pairwise distance matrix for each parameter (with a scaling weight $w$ as a hyperparameter), which generated similarly successful results (not included in the results section, for brevity).}
    \end{itemize}
\end{enumerate}

\subsubsection{Quantization}
No neurons are deleted, instead all parameters in the model are saved as 16-bit floats instead of our default 32-bit floats.

\subsubsection{Trained From Scratch}
 We chose various architectures to train fully from scratch (with random parameter initializations) on the original dataset.
\newline
\newline
First, we trained from scratch (with random weight initialization) the architecture implied by our best (OBD-SD) neuron pruning method applied globally (with a small penalty added to ensure layers don't disappear entirely), which resulted in strongly decreasing layer sizes: 587, 146, 46, 57, 7 that reduced model file size by approximately 75\% (original layer sizes were 1024, 768, 512, 512, 512).
\newline
\newline
We also trained from scratch the various architectures yielded from removing a fixed proportion of neurons from each main fully connected layer, including one that also reduced model file size by about 75\% for comparability. These tended to produce better final results than pruning methods yielded (see Figures 4 and 6).

Finally, we also trained from scratch various architectures yielding from removing a fixed proportion of neuron \textit{connections} (parameter-level pruning) in the first five fully connected layers of our model (analogous to the `random' parameter-level pruning approach). These tended to produce similar final results than our best pruning methods yielded (see Figures 3 and 5).

\section{Results}

\begin{figure}[htbp]
    \centering
        \includegraphics[width=.75\textwidth]{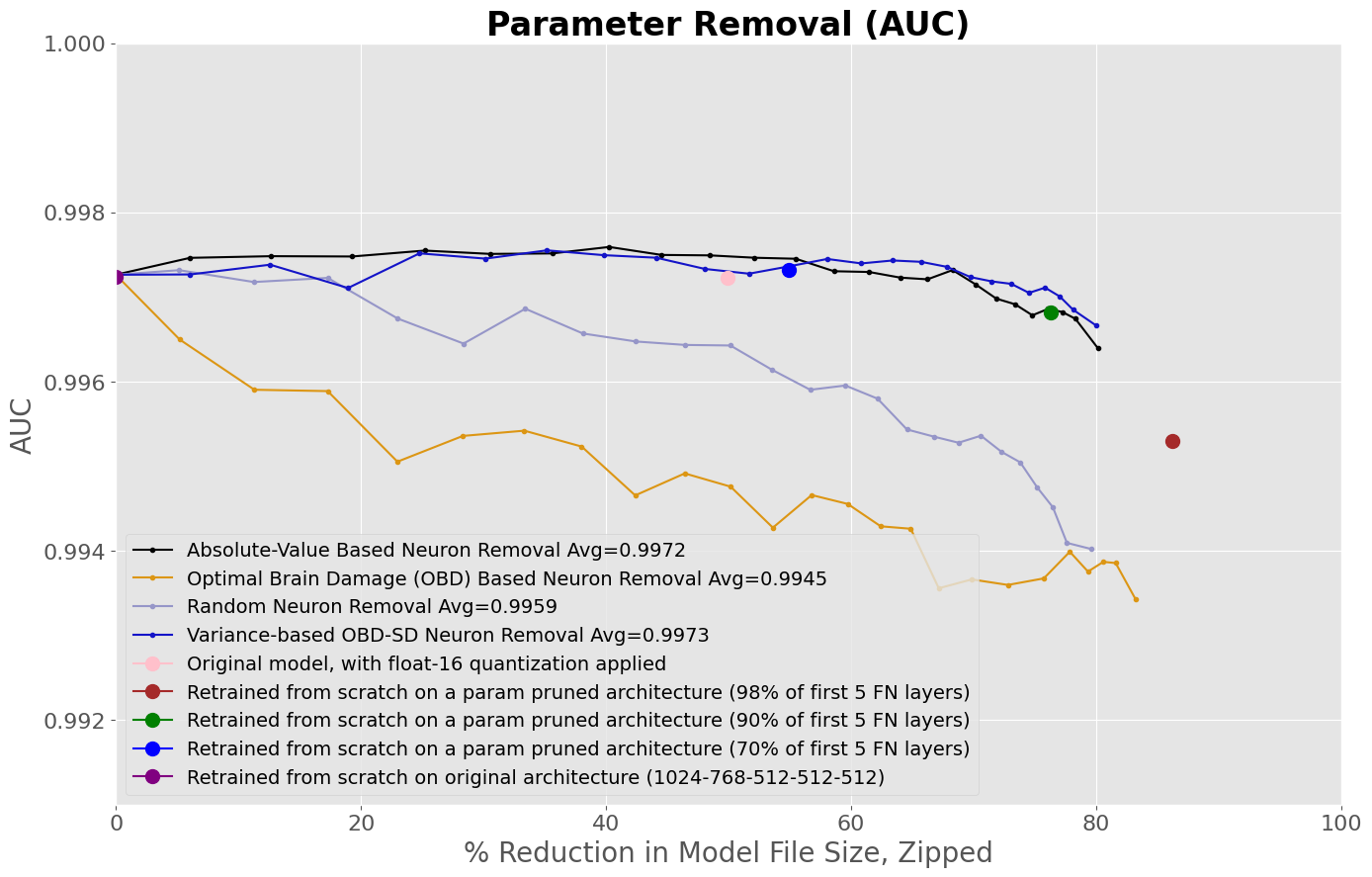}
        \caption{AUC comparison of various parameter pruning methods (using 10\% pruning on fully-connected layers with one fine-tuning epoch each round). Models that underwent random parameter pruning, random initialization, and then a trained fully (20 epochs) are shown as large dots.}
\end{figure}

\begin{figure}[htbp]
    \centering
        \includegraphics[width=.75\textwidth]{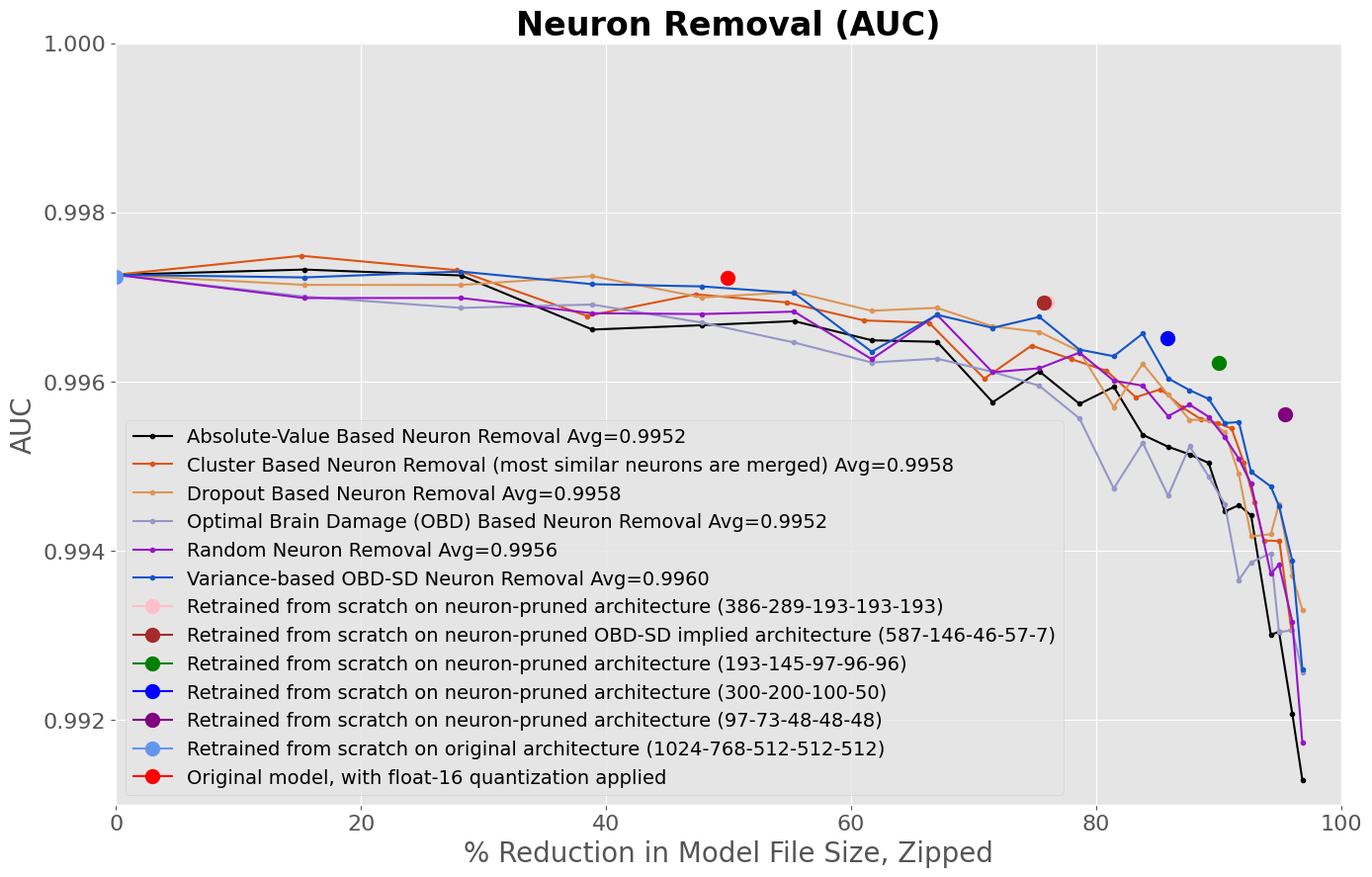}
        \caption{AUC comparison of various neuron pruning methods (using 10\% pruning on fully-connected layers with one fine-tuning epoch each round). Smaller model architectures (analogous to neuron pruning) that underwent random initialization and then were trained fully (20 epochs) are shown as large dots.}
\end{figure}

\begin{figure}[htbp]
    \centering
        \includegraphics[width=.75\textwidth]{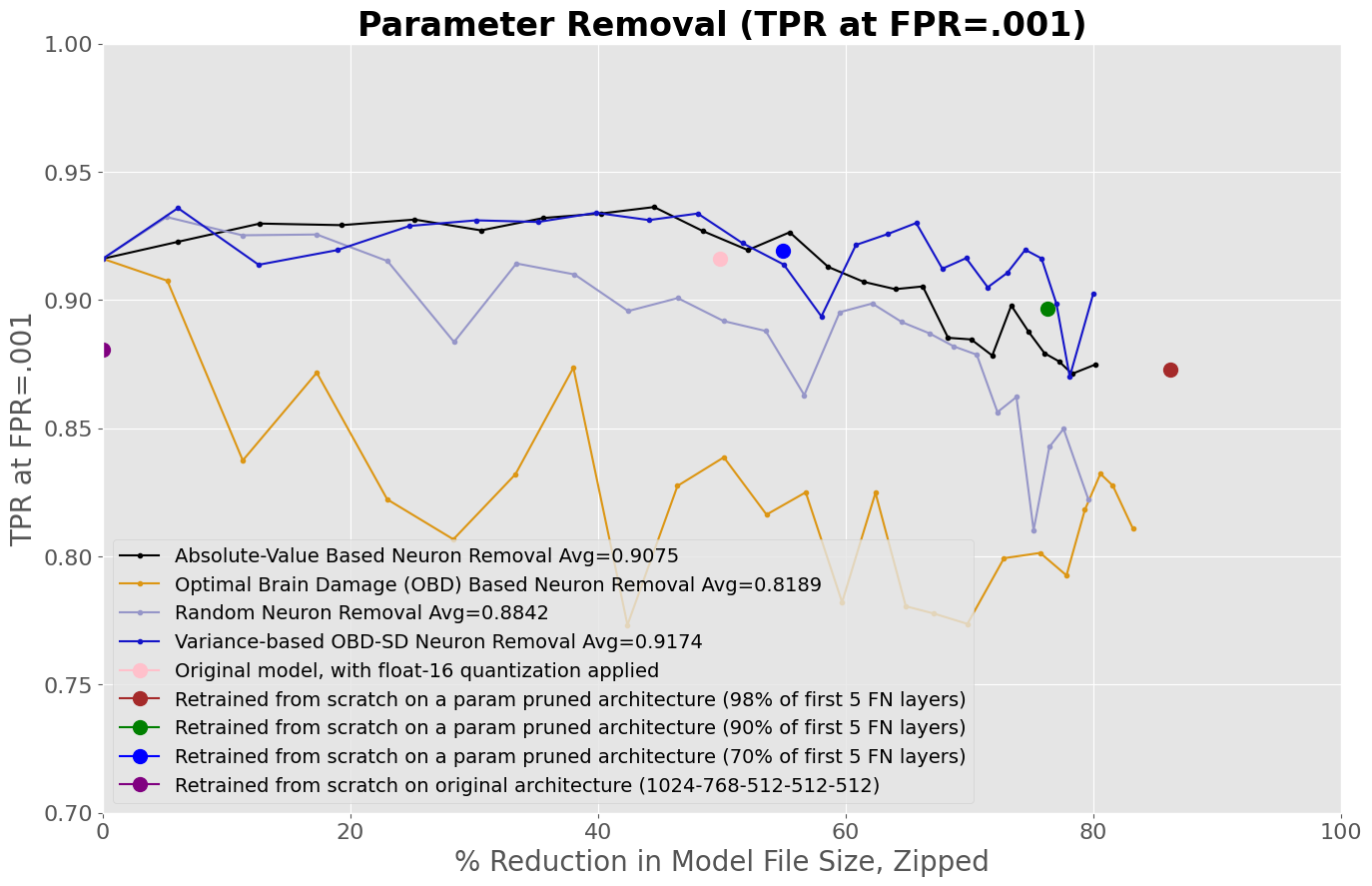}
        \caption{TPR at a FPR of .001 comparison of various parameter pruning methods (using 10\% pruning on fully-connected layers with one fine-tuning epoch each round). Models that underwent random parameter pruning, random initialization, and then a trained fully (20 epochs) are shown as large dots.}
\end{figure}

\begin{figure}[htbp]
    \centering
        \includegraphics[width=.75\textwidth]{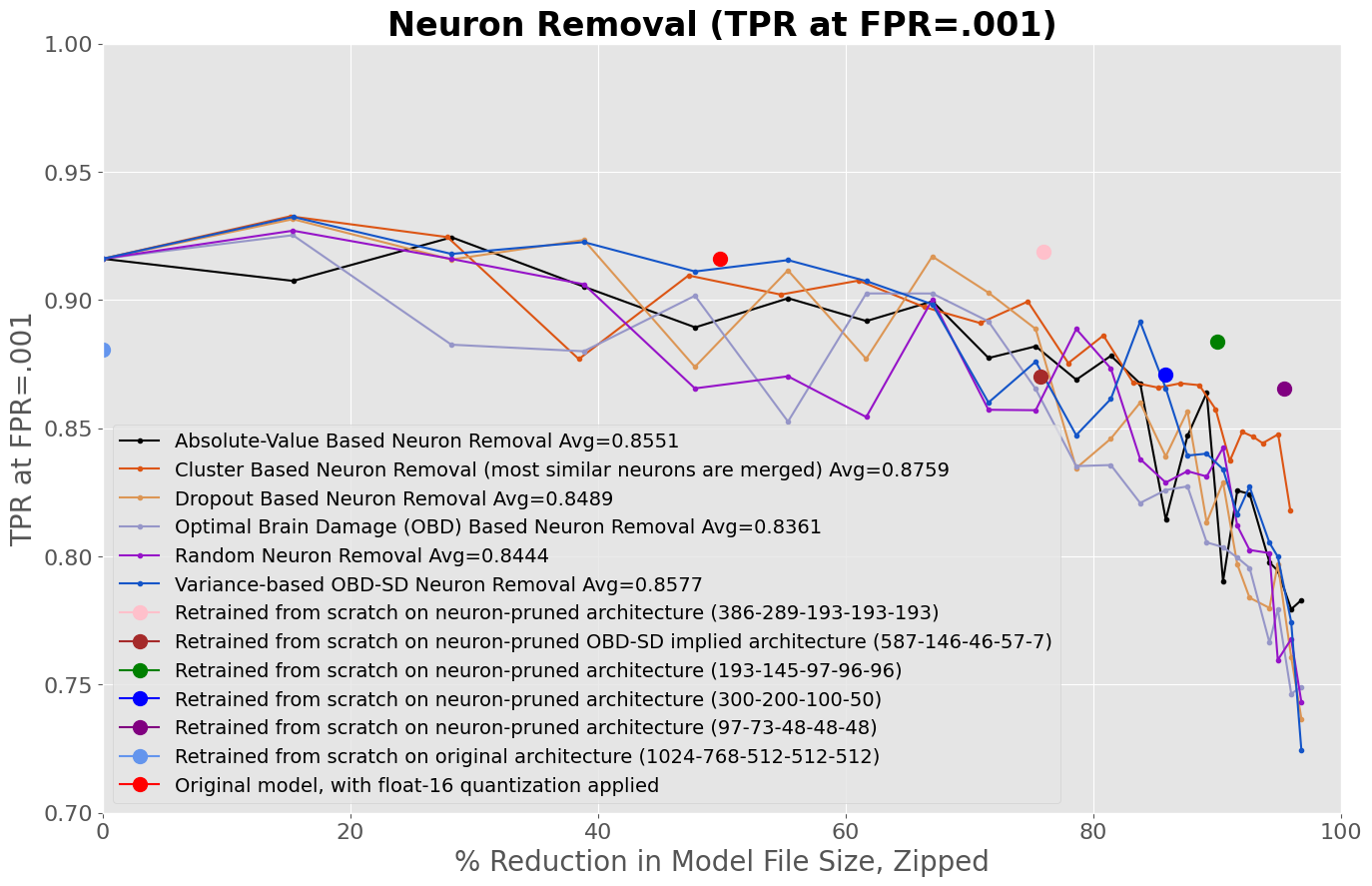}
        \caption{TPR at a FPR of .001 comparison of various neuron pruning methods (using 10\% pruning on fully-connected layers with one fine-tuning epoch each round). Smaller model architectures (analogous to neuron pruning) that underwent random initialization and then were trained fully (20 epochs) are shown as large dots.}
\end{figure}

Interestingly, the OBD approach did consistently worse than random\footnote{You might be wondering if a negative was dropped somewhere in our OBD equations (we did, initially!) - but nope, -OBD damage estimates did even worse than the OBD formula, crucially because (at least in our model) the parameters that have little effect on the model tend to have OBD damage estimates close to 0 (with low sample variance), whereas the OBD damage estimate tails' are more strongly negative and positive - further away from zero. This is due to calculating the second partial derivative directly in our OBD implementation, instead of using an imprecise approximation - discussed more in section 3.1.}. Investigating \textit{why} led us to develop our OBD-SD modified approach (see section 3.1), which appeared to to perform best out of all our pruning comparisons particularly when heavily pruning models (absolute-value based parameter pruning did best for small amounts of pruning).
\newline
\newline
The pruning results are less clear when removing entire neurons at once (although OBD-SD still seemed to come out a bit ahead, especially with large amounts of pruning). This makes sense: a neuron in our neural network is associated with hundreds to thousands of parameters. It's far more difficult to identify a neuron that is unlikely to cause much damage, purely because it's unlikely that all of their parameters are unimportant. So while some methods appear to do slightly better than the others, the benefit is nowhere near as stark as when methods are applied on a parameter level.
\newline
\newline
When choosing which method to apply, it's important to consider in what way you want your model to be 'smaller'. If uncompressed (unzipped) file size and memory are the things you wish to minimize, parameter level pruning isn't particular useful (at least, not without sparse matrix operation code). However, if zipped file size is the thing you wish to minimize (often the case when deploying model to many endpoints), then parameter level pruning is performs significantly better than neuron-level pruning.
\newline
\newline
However, even the winning pruning methods performed only \textit{slightly} better than training a model from scratch on a smaller size. Liu et. al\cite{rethinking} asserts that fine-tuning a pruned model only gives comparable or worse performance than training the same model with randomly initialized weights. Whereas the Lottery Ticket Hypothesis\cite{lottery} indicates that pruning and fine-tuning a model will produce better results. Our results side with the Lottery Ticket Hypothesis, but the margin is slim (possibly `comparable') - and only true for parameter level pruning.
In our results, the most successful parameter-level pruning methods did slightly better than training from scratch, whereas neuron-level pruning did significantly worse than retraining from scratch (likely because parameter-level pruning is able to target irrelevant parameters more precisely).
\newline
\newline
Float-16 Quantization also performed far better than pruning methods (approximately halving model size with barely an affect on accuracy), though is somewhat limited in terms of flexibility.



\section*{Conclusion}
While parameter-level pruning is effective at reducing zipped model size, and neuron-level pruning is effective and reducing overall (in-memory, uncompressed, and zipped) model size, our results suggest that simply training a smaller model from scratch yields the same or better results. Quantization is also a simple and good approach. While our new parameter damage estimation technique, OBD-SD, seems to do about the same or better than all other pruning methods we tested, we conclude that it's not particularly useful to prune with such methods \textit{unless} you want to maintain your ability to easily trade off between accuracy and model size (because pruning generates many models of varying sizes, without having to train each different model from scratch).


\section*{Acknowledgements}
Thanks to Sophos for supporting this research.

\printbibliography

\end{document}